\documentclass[letterpaper, 10 pt, conference]{ieeeconf}
\IEEEoverridecommandlockouts
\usepackage{cite}
\usepackage{amsmath,amssymb,amsfonts}
\usepackage{algorithmic}
\usepackage{graphicx} 
\usepackage{color}
\usepackage{xcolor}
\usepackage{subcaption}
\usepackage{hyperref}
\usepackage{float}
\usepackage{diagbox}
\usepackage{comment}
\usepackage{soul}
\usepackage{tabularx} 

\def\BibTeX{{\rm B\kern-.05em{\sc i\kern-.025em b}\kern-.08em
    T\kern-.1667em\lower.7ex\hbox{E}\kern-.125emX}}
\begin{document}

\title{\LARGE \bf As You Wish:  Mission Planning with Formal Verification using LLMs in Precision Agriculture }

\author{Marcos Abel Zuzu\'{a}rregui \qquad Stefano Carpin%
\thanks{
The authors are with the Department of Computer Science and Engineering, University of California, Merced, CA, USA.
This work is partially supported by the IoT4Ag Engineering Research Center funded by the National Science Foundation (NSF) under NSF Cooperative Agreement Number EEC-1941529 and under grant CMMI-2326310. Any opinions, findings, conclusions, or recommendations expressed in this publication are those of the author(s) and do not necessarily reflect the view of the National Science Foundation.}
}


\maketitle

\begin{abstract} 
Though robotic systems are now being commercialized and deployed in various industries, many of these systems are highly specialized and often require an advanced skill set to operate and ensure they perform as instructed. 
To mitigate this problem, it has been proposed
to use large language models (LLMs)
to synthesize mission plans in precision agriculture and other domains based on mission descriptions provided in natural language (NL).
While these systems demonstrate impressive performance, they also suffer from the inherent ambiguities of NL.
In this paper, we address this issue by introducing
a planning architecture that combines LLMs with linear temporal logic (LTL) to ensure that, through formal verification, the mission planning system meets the specifications formulated by the user while still using NL.
In our proposed  system, the mission plan is seen as
the implementation and the LTL formalization is seen as the specification. 
 Both are automatically extracted from mission descriptions provided in NL.
To mitigate potential bias, two separate LLMs are tasked with the
implementation and specification generation.
Through feedback loops, the system self-corrects when syntax
or verification errors are encountered, thus offering a fully hands-off
solution.
Through extensive experiments, we highlight the strengths and limitations of integrating mission verification into a fully autonomous pipeline, particularly regarding an LLM's ability to generate valuable LTL formulas, and show how our proposed implementation addresses and solves these challenges.


\end{abstract}


\section{Introduction}  
As robot systems find their way into novel industries and sectors, the need for simplified robot interactions continues to grow. 
While in the academic sphere expecting a foundational knowledge of operating robots might be assumed, having the same expectations from non-technical end users prevents widespread technical adoption. 
This is especially true in our domain of interest, precision agriculture, where solutions that make controlling robots easier are in high demand. 
We now see large language models (LLMs) acting as conduits for previously specialized industries to interact with robotics using natural language (NL) \cite{mower_ros-llm_2024, CarpinLLMSubmitted, kannan_smart-llm_2024}.
However, given how ambiguous NL can be, \textit{how can
we ensure that our systems are doing what we intend them to do}? 
This question serves to highlight the difficulties faced in generic robotic mission planning (MP) systems. 
This problem is due to the objective ambiguity of NL used in everyday conversations. 
Therefore, in many situations one is not really sure about how the system has interpreted the assigned mission or is faced with often tedious requests for confirmation of the interpretation before execution commences.
These difficulties are magnified when accounting for situations in which the outcome is stochastic and dependent on factors only uncovered during runtime. 
An additional challenge peculiar to precision agriculture is that robots in rural environments often operate without network connectivity, preventing real-time access to remote services, and can be potentially dangerous due to the size of the autonomous vehicles involved in agricultural operations.

\begin{figure}[tb!]
\centering
\includegraphics[width=0.9\linewidth]{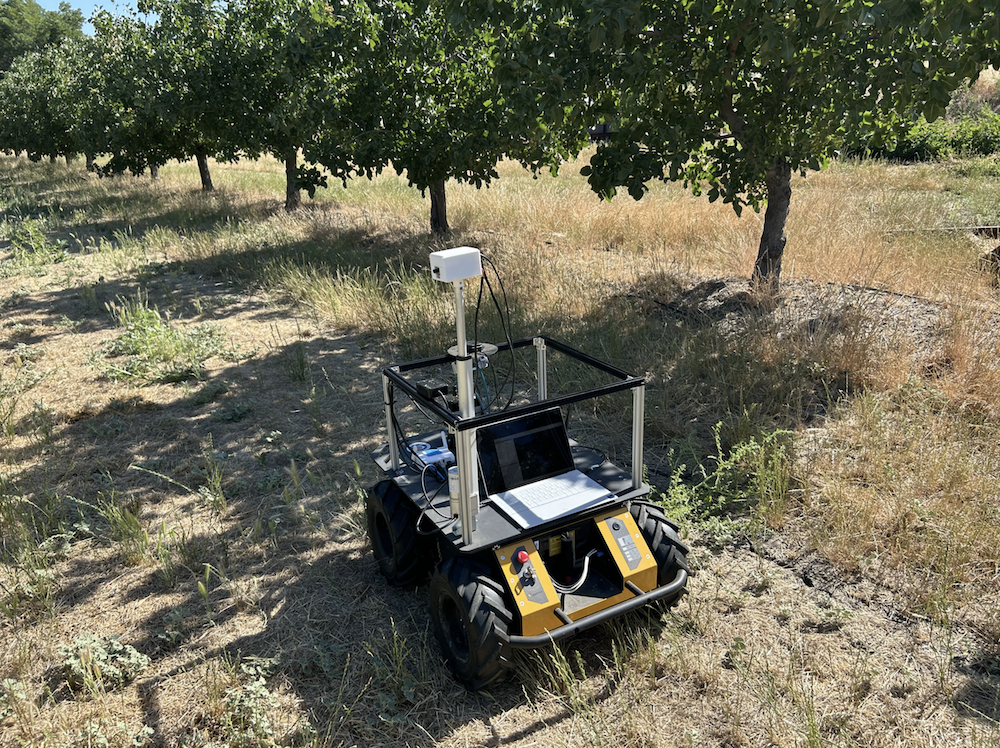}
\caption{A task in precision agriculture: a robot acquiring thermal imaging
of a pistachio tree for water stress detection. }
\label{fig:graph20_ucm}
\vspace{-3mm}
\end{figure}

Since agricultural robots are intended to be used by non-specialists, it is essential to enable users to create complex mission plans without needing to manage low-level details or grasp system-level complexities.
In \cite{CarpinLLMSubmitted}, the authors present an open-source LLM-based mission planner that automatically translates NL into robot missions that can be executed by a mobile robot. 
In our tests with this planner, we observed that, occasionally, the system produces and executes missions that do not necessarily align with user intent. 
We posit that this problem can be addressed as an instance of \emph{formal verification} (FV), where one is concerned with verifying whether a given implementation satisfies an assigned specification \cite{BaierBook}. 
In this case, the implementation is the mission plan synthesized by the LLM \cite{CarpinLLMSubmitted}, while the specification is the mission description provided by the user in NL. 
While there is a rich literature on FV, one stumbling block is that the tools used to produce specifications, like linear temporal logic (LTL), cannot be assumed to be known by the casual user. 
Therefore, we build upon the open-source system presented in \cite{CarpinLLMSubmitted} and designed a separate LLM-based pipeline with
the ability to automatically synthesize mission specifications in LTL solely from NL user input.
Additionally, the system automatically checks the original mission plan against the LTL specification using the open-source SPIN model checker \cite{SPINBook}, and when discrepancies are identified, it autonomously feeds back error messages to the system for self-correction.
All of these functionalities are achieved without any additional input required to the end-user, thus  lowering 
the learning curve to enable non-technical users to plan for missions with confidence and offering a fully hands-off solution.


To achieve our stated goals, we use two independent AI agents, in a single framework, with the express goal of guaranteeing a mission plan that matches a given NL prompt. 
The two agents are responsible for robotic task representation, defined by the IEEE standard \cite{noauthor_ieee_2024}, and LTL generation to validate the intended task sequence, respectively. 
The contributions of this paper are the following:
\begin{itemize}
    \item we present an LLM-to-robot task execution pipeline that is formally verified for autonomous navigation and data collection; 
    \item we show how mission plans generated leveraging
    LLMs can effectively be verified also using LLMs without a human in the loop;
    \item we validate our proposed system with real end-users and show limits and strengths.
\end{itemize}

The rest of the paper is as follows. 
Selected related work is presented in Section \ref{sec:sota}. 
In Section \ref{sec:method}  we describe the system we developed, with experiments detailing our findings are given in Section \ref{sec:results}. Finally, conclusions and future work are presented in Section \ref{sec:conclusions}.
\section{Related Literature} \label{sec:sota}
\subsubsection{LLMs for Mission Planning}
With the increasing adoption of LLMs across various domains, MP has been no exception. 
However, LLMs have a documented history of being undependable, at times \cite{emsley_chatgpt_2023, ray_chatgpt_2023}.
We ask the question, \textit{how can we integrate the general power of LLMs in MP with any form of certainty}?
Recent literature has explored the integration of LLMs in manipulator planning with executable code \cite{mower_ros-llm_2024, kannan_smart-llm_2024, liang_code_2023, cheng_empowering_2024}, as well as in mobile robotics \cite{CarpinLLMSubmitted}, demonstrating the growing interest in leveraging LLMs for MP. 
While these approaches share the overarching objective of facilitating user-friendly interaction without compromising performance, this work specifically addresses plan verification in the absence of human oversight. 
Existing LLM-based planners all face this critical limitation in their ability to operate independently and autonomously.

In prior studies, mission plans have been validated using formal methods such as Planning Domain Definition Language (PDDL) and Linear Temporal Logic (LTL) \cite{kalluraya_multi-robot_2023, kalluraya_resilient_2023, noauthor_ieee_2024}. 
However, utilizing these methods typically requires users to provide custom inputs, necessitating familiarity with PDDL or LTL plus human intervention.
In contrast, the proposed architecture eliminates the need for user expertise in formal planning languages by employing an LLM to autonomously generate LTL specifications. 
Verification is then conducted by an independent, uninformed third-party LLM, effectively mirroring the role of human oversight in mission validation.

\subsubsection{LTL}
As mentioned, LTL has been used in mission planning \cite{kalluraya_multi-robot_2023, kalluraya_resilient_2023}, but is used as the \textit{input} to the verification system mission rather than a complement to the data flow.
In literature outside of planning, LLMs have been used to generate LTL
formulas \cite{pan_data-efficient_2023, murphy_guiding_2024, li_automatic_2025} or understand formal software verification \cite{jansen_can_2023}.
SPIN \cite{SPINBook}, a model checking tool, provides the foundation for exploring system model state spaces.
This can be paired with LTL verification.
Spot \cite{duret.22.cav}, a platform for LTL and automata processing, represents LTL forumlae in memory via B\"uchi automata.
These two tools are used offline to evaluate a system model and its expected behavior.
However, we leverage them as online tools to not only validate a given model, but also provide syntax checking and generate accepting runs.
To our knowledge, this paper is the first to attempt to formally validate LLM mission plans by automating an LTL generator in parallel, creating fully automated plans with associated system verification.

\section{System Architecture and Design} \label{sec:method}
\begin{figure*}[htb]
\centering
\includegraphics[width=0.9\linewidth]{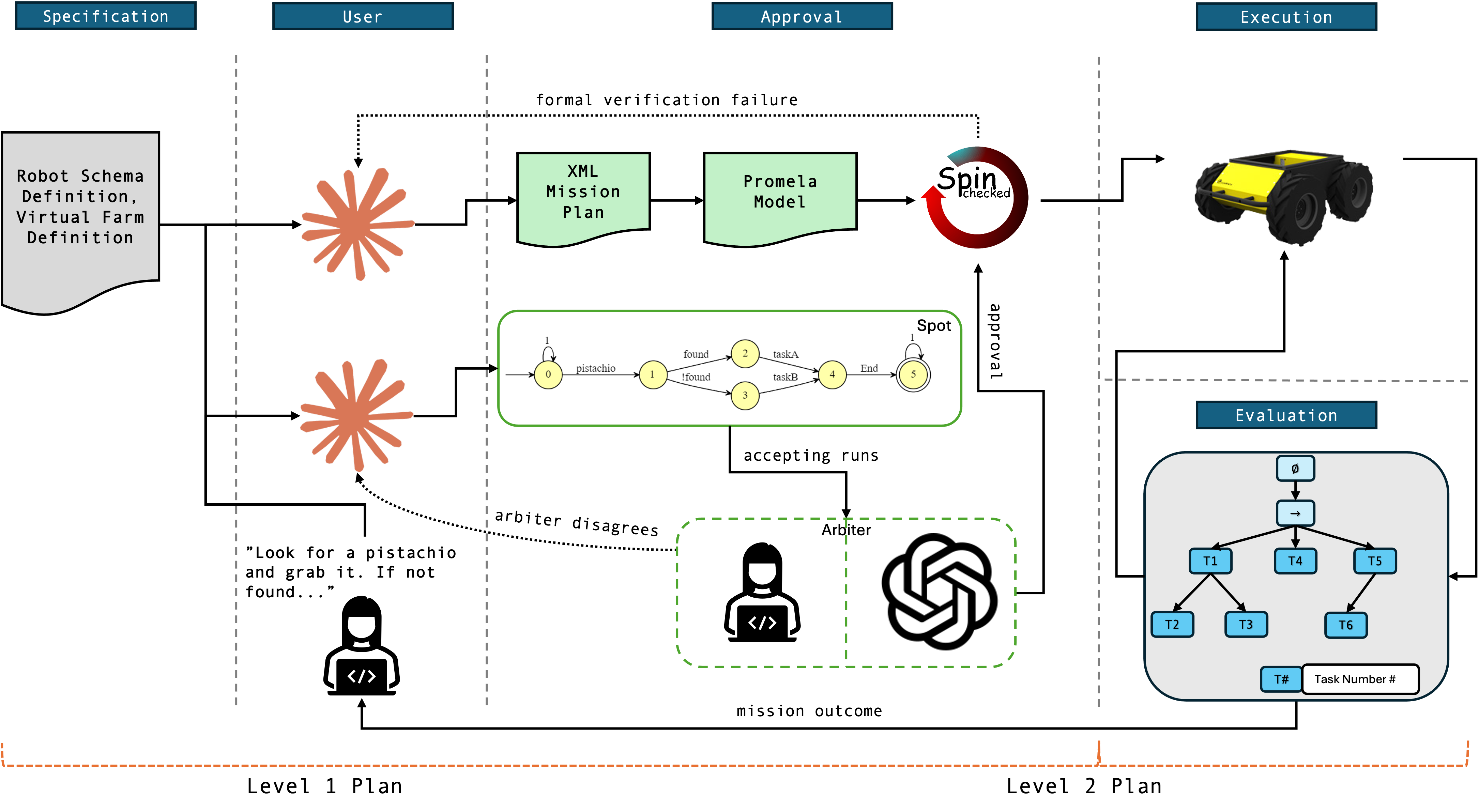}
\caption{Our proposed architecture builds upon \cite{CarpinLLMSubmitted} and follows the guidance of \cite{noauthor_ieee_2024}. It breaks up MP into five functional roles. The focus of this paper is in the \textit{approval} stage (figure adapted from \cite{CarpinLLMSubmitted}).}
\label{fig:spin-architecture}
\end{figure*}

Starting from the open-source mission planner \cite{CarpinLLMSubmitted}, 
we improve it by introducing a novel solution to robustly and autonomously verify the correctness of NL-based mission plans.
{\em Correct}, in this context, means that the mission executed matches the user intent.
Our key contribution is an  architecture that decouples implementation and formal verification to  ensure that the mission specified is carried out as intended by the user.
We refer the reader to \cite{CarpinLLMSubmitted} for a deeper discussion of the original system we are extending and to understand the IEEE framework proposed in \cite{noauthor_ieee_2024}.
Our proposed system is best visualized in Figure \ref{fig:spin-architecture} which outlines the two-level design showing the separation from the high-level planning and the low-level execution. 
Within the two levels, we find five stages (specification, user, approval, execution, and evaluation) that will be individually described in the remainder of this section.
We start with a brief formalization of the mission planning problem in Section \ref{sec:problem} before moving to system design starting with Section \ref{sec:spec}.

\subsection {Problem Definition} \label{sec:problem}

Robot mission planning has been defined in multiple ways in the literature. We adopt the classic discrete feasible planning formulation (see \cite[chap.~2]{LaValleBook}), which comprises a non-empty state space \( S \), an action space \( A \), a transition function\footnote{For brevity, we consider the case where each action \( a \in A \) can be applied in any state in \( S \), though the formulation can be easily extended to allow state-specific action sets. This extension would introduce the classic concept of \emph{preconditions}.}  
\[
T: S \times A \rightarrow S
\]  
an initial state \( s_0 \in S \), and a set of goal states \( S_G \subset S \).  
In the feasible planning problem, the objective is to find a sequence of actions that, when applied in order, transition the system from the initial state \( s_0 \) to any state in \( S_G \). Traditionally, one would provide an explicit or implicit representation of \( S, A, T, s_0 \), and \( S_G \), and then use a search algorithm to explore the associated search graph and determine whether a solution exists.  
In this architecture, the action set \( A \) 
is defined as a collection of capabilities encoded as ROS2 \emph{actions}, each corresponding to a specific robot platform. The key innovation is that once this action pool is properly represented and provided to the LLM as context, the LLM infers the remaining components:  
\[
S, s_0, T, \text{ and } S_G.
\]  
As elaborated later, this inference process draws upon the full context and the user's query.  
One drawback, addressed in this work, is that the inference is prone to the ambiguities 
of natural language and the plan may result in executions that do not match user intent.

\subsection{Specification} \label{sec:spec}
The \textit{specification} stage provides critical information for the decomposition of the Level 1 (L1) plan. 
A L1 plan defines a task sequence of a high-level mission without defining the details of how it is done, as per the IEEE standard \cite {noauthor_ieee_2024}.
At this stage, relevant context files are provided. 
These context files may include various forms of world information and robot specifications; however, they must always contain an XML Schema Definition (XSD) file that defines the robot’s capabilities -- corresponding to the action set $A$ discussed in Section \ref{sec:problem}.
These context files are what the system will use to decompose a L1 plan, so anything relevant must be included.
Beyond specifying the action set, the XSD file imposes constraints on the resulting XML mission plan, ensuring that it conforms to a behavior tree structure. 
This constraint simplifies the decomposition of the L1 plan at the robot level. 
The XSD adheres to the standard proposed in \cite{noauthor_ieee_2024}, enabling seamless integration of new robots by following a standardized methodology.
The robot's capabilities, referred to as atomic actions, are defined within the XSD and represent the only components requiring modification when introducing a new robot with a different set of capabilities -- i.e., the available actions for the planner. 
These modifications primarily involve updating the XSD to include new robot action definitions, their corresponding parameter specifications, and other relevant state information.
Finally, a feature introduced in our implementation, we provide a GPS polygon of the farm along with an orchard layout, which may be represented as a regular grid (e.g., 10 $\times$ 10 trees) or as an irregular arrangement.


\begin{figure*}[tb!]
\centering
\includegraphics[width=0.8\linewidth]{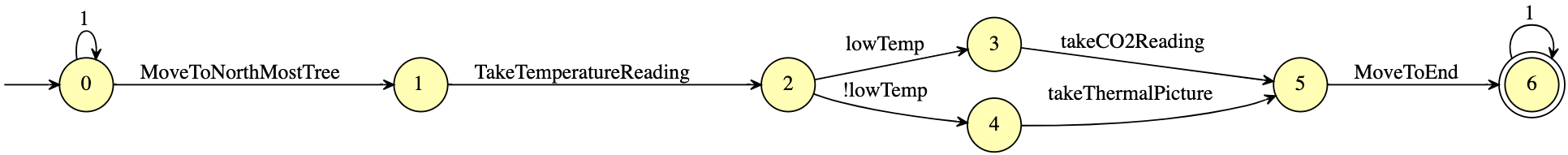}
\caption{Example B\"{u}chi automata of the LTL formula given at the end of Section \ref{sec:cosafe}.}
\label{fig:buchi}
\end{figure*}

\subsection{User} \label{sec:user}
In \cite{CarpinLLMSubmitted}, there was only one LLM agent responsible for generating an XML mission plan.
The limitation of this approach is that a single agent can ensure that the produced files encoding the mission are syntactically correct but cannot guarantee that the produced mission is aligned with the user intent.
In this paper, we introduce another LLM agent responsible for producing a LTL specification to verify the XML mission produced by the first agent.
It is important to note that these two agents are completely decoupled and do not share information at their initial mission generation stage to remove bias.
In the \textit{approval} stage, they will exchange syntactic information to recouple the previously independent code after confirming a fundamental agreement in mission decomposition.
We will discuss the implications of this design in Section \ref{sec:approval}.
The LTL agent creates a LTL formula that is verified to compile in SPIN. 
These LTLs must conform to a LTL subclass known as \emph{co-safe} LTL formulas \cite{he_towards_2015}, which have proven to be valuable for robot mission planning.
More details on this aspect are given in Sections \ref{sec:approval} and \ref{sec:cosafe}.  
We use Anthropic's Claude, see Section \ref{sec:results}, for both agents to decompose the mission plan for their respective duties.
Specifically, the LLMs analyze the provided mission space to generate the remaining components of the planning problem.
In the case of the XML mission generating agent, $S,T,s_0$ and $S_G$ are produced.
In the case of the verification agent, a LTL formula compatible with SPIN is produced.

\subsection{Approval}\label{sec:approval}
The \textit{approval} phase is the critical step to ensure that the mission plan generated by the LLM is valid and matches user intent.
We update the previous design of \cite{CarpinLLMSubmitted} to not only have syntactic checking via the XML linter but also formal verification of a system model using the SPIN model checker.
As referenced in Section \ref{sec:user}, there is an initial handshake that the two LLMs must have prior to proceeding.
To eliminate bias by having the agents talk to each other, which would defeat the purpose of the verification, the system first asks for the agents to simply agree on how many tasks there are in the mission. 
This step can easily be expanded to be more sophisticated, but we noted through experiments that 
sharing more information between LLMs introduces bias even with the most trivial information.
With respect to generating the L1 mission plan, this design takes the previously implemented XML generation step from our LLM agent and then runs the output through our custom
designed XML to Promela\footnote{Promela is the process meta language needed by the SPIN model checker to describe the system implementation.} converter.
This converter translates  a mission coded in XML  into the language Promela. 
This step produces a system model compatible with SPIN 
and is guaranteed to be feasible due to the previous lint checking of the XML mission.

Newly introduced, we add another LLM agent to produce a SPIN compliant LTL formula capturing the intent of the NL mission description provided by the user.
The goal is to check if a LTL formula generated from the validation agent is satisfied by the mission generated by the XML agent. 
This step, however, introduces an additional challenge, i.e.,  determining
if the LTL formula generated is consistent with the NL specification, as the LLM provides
 no such guarantee.
To tackle this challenge, we use use Spot \cite{duret.22.cav},
a software tool that takes the generated LTL and converts it into a B\"{u}chi automata that can then be used by
Spot to generate accepting runs.
At this point, the system can be configured in one of two ways: human-in-the-loop or automated arbiter.
Note that for the experiments carried out in Section \ref{sec:results}, we assume the automated arbiter.
In either case, Spot is used to generate a configurable number of accepting runs to demonstrate possible traces through the automata it generated from the LTL formula. 
With these finite traces, the system prompts the arbiter to ask if the example traces satisfy the user's idea of the mission.

In the automated arbiter case, we choose a different LLM provider as the arbiter to avoid possible bias emerging when both the generator and the arbiter are from the same provider.
Regardless of which arbiter is queried, the system implements a feedback loop to retry LTL generation should the
arbiter reject the accepting runs as being not aligned with user intent.
When this happens, the system makes the assumption that only LTL generation was faulty if accepting runs could not be approved, though in principle it could also be that the arbiter made a faulty 
judgment.
Once the arbiter accepts the example runs, the final stage of verification is to run the Promela system model and the LTL through SPIN.
At this point, SPIN goes through the state space of the system and verifies if  the LTL is  violated.
Should a violation occur, with either a human or LLM having approved the LTL, the system now knows that there is a fault in the implementation, i.e., the mission generated by the LLM is not 
aligned with user intent, as determined by the arbiter.
This information is then shared back to the XML generation LLM for another attempt at mission generation.
Once Spin determines that the implementation satisfies the LTL specification, 
the XML mission is shared to the robot to begin execution. 

\subsection{Execution and Evaluation}
Barring a final XML lint to ensure successful plan receipt, \textit{execution} begins by decomposing the XML L1 input into a Level 2 (L2) plan.
This is accomplished with a compiler that parses XML into a behavior tree structure.
Since we know the structure of the XML mission, we can guarantee a feasible behavior tree based on the L1 plan in whatever framework the robot supports.
Each task in the tree is mapped to a software module that will handle the functionality, along with the associated parameters.
Together with the \textit{evaluation} stage, the behavior tree awaits task outcomes to assess mission progress or behavior tree path selection. 
Our improved system continues to have modularity and flexibility, by design, to augment missions as a whole or by task.
This flexibility was demonstrated in \cite{CarpinLLMSubmitted} augmenting a general L1 plan to fit execution constraints not explicitly stated in the mission.

\subsection{Co-Safe LTL} \label{sec:cosafe}
Introduced in Section \ref{sec:approval}, LTL is a formalism to describe properties that the implementation must satisfy.
In dealing with LTL and formal verification, it is important to understand how these algorithms work to quantitatively measure success. 
SPIN will take the Promela system model, compiled from XML, and validate against the LTL.
This validation is performed through  an exhaustive  state space exploration algorithm, ensuring that no feasible run can violate the LTL assertion. 
Co-safe LTL \cite{he_towards_2015}, a fragment of LTL that combines Boolean and logical operators
that has proven to be particularly useful in robotics for their inherent ability to describe tasks that are completable in a finite amount of time, a common requirement in real-world robotic applications \cite{CoSafeLTLShoukry}.
Following the notation introduced in \cite{he_towards_2015}, if $\Pi$ is a set of atomic propositions and $\pi \in \Pi$ is a generic atomic proposition, a co-safe LTL formula over $\Pi$ can be recursively defined as follows:
\[
\varphi = \pi | \neg \pi | \varphi \wedge \varphi | \varphi \vee \varphi |
\varphi \mathbf{U} \varphi |  \mathbf{X} \varphi |  \mathbf{F} \varphi
\]
where $\neg, \vee, \wedge$ are the classic logic operators, and
$\mathbf{U},\mathbf{X},\mathbf{F}$ are the LTL operators \emph{until},
\emph{next}, and \emph{finally}.
The reason to consider co-safe LTLs  formulas is that   they can be verified in finite time by checking a finite length prefix, thus aligning with the process we outlined where Spot generates
finite lenght runs that are passed to the arbibter for approval or rejection.
The classic LTL definition includes temporal operators such as \textit{always} that can require prefixes of infinite length to be verified or violated.
Co-safe LTLs eliminate the use of these operators, forcing the LTL to ensure liveness of the system.
To exemplify the functionalities described thus far, 
consider the following mission description expressed in NL, 
"\textit{Move to the north most tree and take a temperature reading. If lower than 30$^\circ$C, take a $CO_2$ reading. If not, take thermal picture.}"
The formula shown in Eq.~\eqref{eq:ltl} shows the LTL generated by the LLM during verification, while
 Figure \ref{fig:buchi} displays the corresponding automaton generated by Spot.
\begin{flalign}
    & (\text{moveToNorthMostTree} \, \land \mathbf{X} (\text{takeTempReading} \, \land  \notag \\
    & \mathbf{X} (
        ((temp < 30) \land 
            \mathbf{X} \, \text{takeCO2Reading}) \, \lor  \notag \\
    & ((temp \geq 30) \land 
            \mathbf{X} \, \text{takeThermalPic})))) & \label{eq:ltl}
\end{flalign}

\subsection{Data Flow Example}\label{sec:example}
\begin{figure}[tb]
\centering
\fontsize{6}{10}\selectfont
\includegraphics[width=0.5\textwidth]{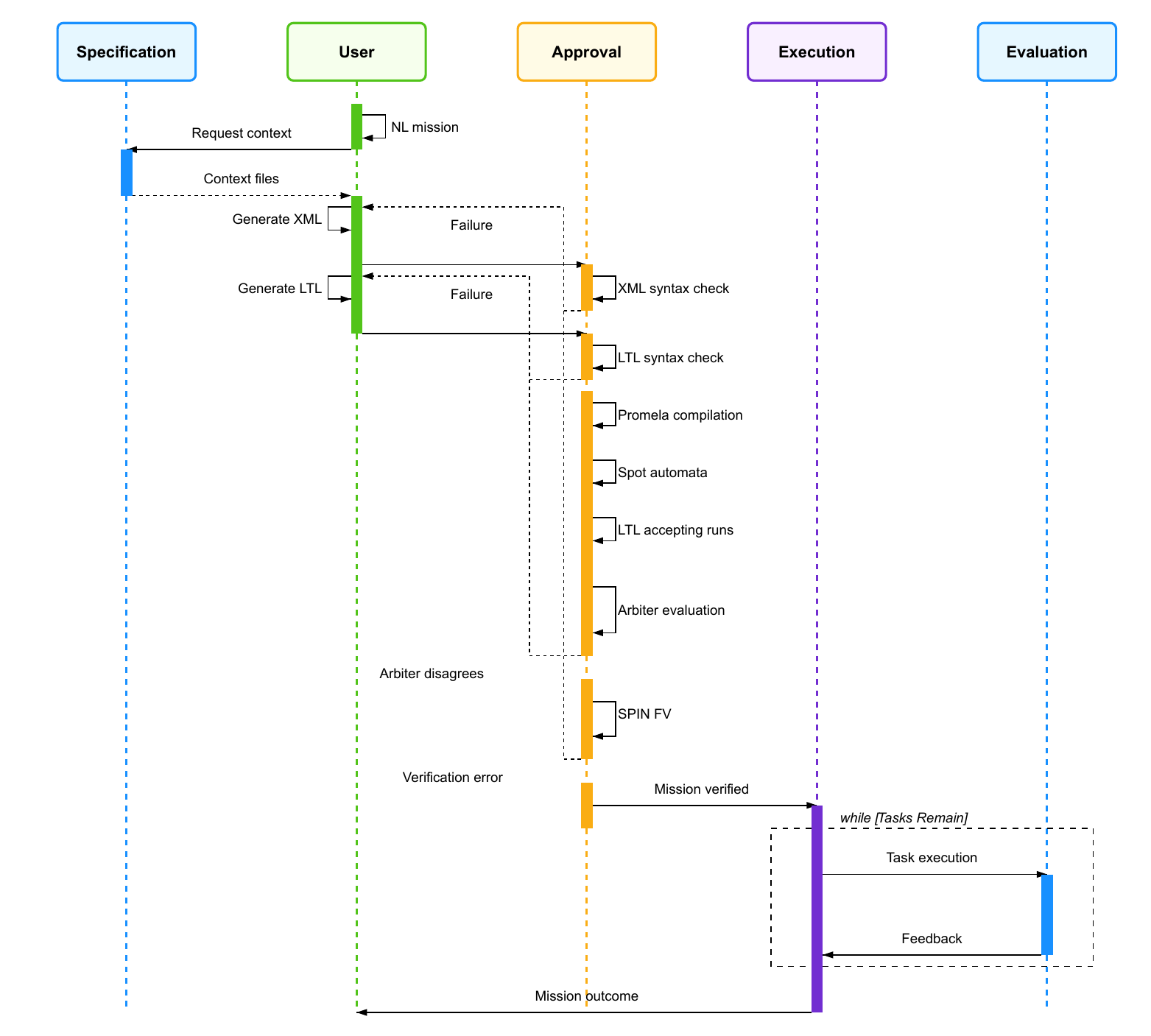}
\caption{Sequence Diagram for Section \ref{sec:example} example query.}
\label{fig:sequence}
\end{figure}

In this section we present an example mission data flow below to help demystify the intricacies of the system. 
To follow along with this example, we refer the reader to Figure \ref{fig:sequence} as reference.
Assuming context has been prepared in the form of an XSD and farm GPS polygon, the \textit{user} asks the mission presented at the end of section Section \ref{sec:cosafe}.
The system gathers all relevant context files and packages them with this prompt to be sent to the XML and LTL agents, separately.
Each LLM agent then generates its respective task: an XML mission and a LTL mission specification, 
in that order.
Next, SPIN  and Spot  evaluate LTL syntax before generating an automaton. 

Note that if in the current stage there is either a failure with XML or LTL syntax,  
a feedback error message is sent back to the respective agent with a request to address it.
The XML mission is then sent to a Promela compiler that decodes XML to Promela. 
With the automaton from Figure \ref{fig:buchi}, the system generates a configurable number of accepting runs.

One such example run is, "\textit{MoveToNorthMostTree, TakeTemperatureReading, lowTemp, takeCO2Reading}".
These runs are sent to the arbiter (LLM or human), which determines if they align with the mission intent in all cases.
The arbiter has no more information other than the accepting runs, the original NL mission text, and the associated context.
If the arbiter approves, SPIN formally verifies whether the mission (i.e., the implementation)
satisfies the specification provided by the LTL formula.
If SPIN determines a violation, due to arbiter approval the system assumes that the 
implementation is not correct, and the XML mission should be reworked accordingly.
With a verified mission, the L1 mission is sent over TCP to be decomposed into a L2 plan for the robot to 
execute. 
The user is also referred to the companion video for 
a visual representation of the the data flow through the
feedback loops.

\section{Results} \label{sec:results}

In this section, we show how system can formally verify missions in precision agriculture settings.
Due to space limitations, only a subset of results are presented. 
The remaining queries and relevant information can be found 
on the anonymized website \url{https://ucmercedrobotics.github.io/asUwish.html}.
Given the focus of this paper is on formal verification and not mission execution, we only evaluated L1 output from the mission planners and executed in Gazebo with ROS2. 
This setup is sufficient to establish if the stated objectives have been achieved or not.

For both XML and LTL generation agents, we used \textit{claude-sonnet-4-20250514} with a temperature of $0.2$.
For the arbiter LLM that validates LTL runs against the desired mission, we used \textit{gpt-5-2025-08-07}. 
Max response tokens for all three models is 10,000.
The total number of feedback loop retries, requerying the LLM with an error message, was 10 throughout all experiments.
This is applied globally, per mission, and not to a single LLM.
Finally, we use SPIN version 6.5.1 and Python Spot version 2.12.2. All code
is freely available under the Apache 2.0 license at the
aforementioned website.

\subsection{Verifying MP Problems} \label{sec:exp}
In experimenting with verification, we start with the mission prompts given in \cite{CarpinLLMSubmitted} to assess whether the extensions we introduced are backwards compatible. 
We divide the previous prompts into two distinct categories: \textit{Explicit} and \textit{Implicit}.
Many of the existing queries were worded explicitly, quantitative by nature.
We then convert some of the previously explicitly worded queries from \cite{CarpinLLMSubmitted} to be implicitly worded, more qualitative by nature, to evaluate how the system performs on the same mission written in two different manners. For example, ``less than 30$^\circ$C'' (quantitative) 
can be restated as ``low" (qualitative).
Lastly, we experiment with prompts directly from our end users such as
farm workers and farm managers. 
Explained in Section \ref{sec:user_study} and categorized as \textit{Farmer} in Table \ref{table:results}, we show example missions obtained from farmers to test the strength of our system on real world missions.
Each of these queries presents their own challenge to the architecture, some of which will be covered in this section.
Results are found in Table \ref{table:results}.
Table \ref{table:results} can be understood by its columns.
\textit{Mission Queries} are shortened representations of the full mission prompts.
Complete prompts can be found on the website for those interested.
\textit{Tasks} are the number of atomic tasks and the number of conditional branches. 
An example of a conditional branch would be validating that a sensor measurement is below or above a certain value as requested by the mission.
\textit{Success} is the number of times all three of the LLMs (XML, LTL, arbiter) agree on the output, SPIN formally verifies the mission, and the user's intent is satisfied.
\textit{Conflicts} begin to be counted 
for errors emerging after
both LLM agents complete their generation tasks and agree on the initial number of tasks. 
At this point should the arbiter decide that the LTL constraints do not satisfy the NL mission or SPIN fails to formally verify the XML model, this is counted as a conflict.
\textit{Recoveries} count if the system was able to resolve a mission conflict on its own and produce the desired mission.
These two columns are quantified as how many conflicts detected by the system were able to be reformulated and subsequently planned correctly.
Possible reasons to reject a mission include misunderstood or ambiguous NL, spatial inconsistencies, overly complex missions, among others.
Even with certain failures, discussed at length in Section \ref{sec:limitations}, the architecture showcases its general flexibility with different types of queries and manages to not only generate relevant L1 mission plans, but formally verify them.

\begin{table}[t]
\centering
\setlength{\tabcolsep}{4pt} 
\begin{tabular*}{\columnwidth}{|>{\centering\arraybackslash}p{0.4\columnwidth}|c|c|c|c|}
\cline{1-5}
\scriptsize \textbf{Mission Queries} & 
\scriptsize \textbf{Tasks} & 
\scriptsize \textbf{Success} & 
\scriptsize \textbf{Conflicts} & 
\scriptsize \textbf{Recoveries} \\
\cline{1-5}
\textbf{\scriptsize Explicit} & & & & \\
\cline{1-5}
\scriptsize ``\textit{4 trees, 2 sensors}" & 8 & 100\% & 0 & 0 \\
\scriptsize ``\textit{Relative conditionals}" & 8 & 100\% & 1 & 1 \\
\scriptsize ``\textit{Relative + absolute}" & 12 & 100\% & 2 & 2\\ 
\scriptsize ``\textit{5 nested if conditionals}" & 13 & 80\% & 3 & 2 \\
\scriptsize ``\textit{If-else with nesting}" & 15 & 100\% & 0 & 0 \\
\cline{1-5}
\textbf{\scriptsize Implict} & & & & \\
\cline{1-5}
\scriptsize ``\textit{North, center, east samples}" & 6 & 40\% & 5 & 2 \\
\scriptsize ``\textit{4 corners relative}" & 8 & 100\% & 0 & 0 \\
\scriptsize ``\textit{Relative + absolute}"$^\dagger$ & 12 & 100\% & 3 & 3 \\ 
\scriptsize ``\textit{5 nested if conditionals}"$^\dagger$ & 13 & 100\% & 0 & 0 \\
\scriptsize ``\textit{If-else with nesting}"$^\dagger$ & 15 & 40\% & 2 & 0\\
\cline{1-5}
\textbf{\scriptsize Farmer} & & & & \\
\cline{1-5}
\scriptsize ``\textit{Plant a row of bittermelon}" & 6 & 100\% & 0 & 0 \\
\scriptsize ``\textit{40 seeds, one yard apart.}" & 6 & 100\% & 0 & 0 \\
\scriptsize ``\textit{Plant 300', skip, plant}" & 13 & 100\% & 0 & 0 \\
\scriptsize ``\textit{2 Chinese bittermelon, 1 Indian}" & 19 & 100\% & 0 & 0 \\
\cline{1-5}
\end{tabular*}
\caption{Scores of various mission prompts executed through the system. All trials are out of 5 attempts each. $^\dagger$ represents implicitly reworded missions.}
\label{table:results}
\vspace{-3mm}
\end{table}

\begin{figure*}[htb]
\centering
\includegraphics[width=1\linewidth]{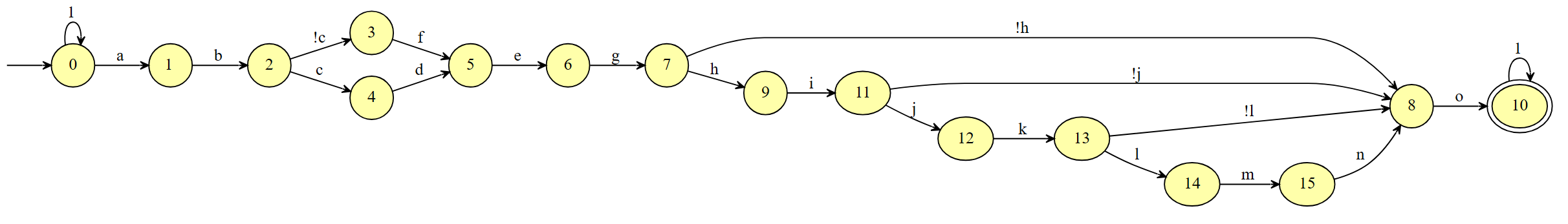}
\caption{Automata generated for the mission, ``\textit{Take a picture of the first tree. If the reading is below 30$^\circ$C, take a CO$_2$ sample. Otherwise, take a temperature reading. Then, measure another tree, a different one by taking a picture. If it's less than 30$^\circ$C, sample CO$_2$. If that reading is less than 400ppm, sample CO$_2$ again to confirm. If still less than 400, continue driving to another tree and take another CO$_2$ sample. finally go to end.}"}
\label{fig:complex_aut}
\end{figure*}

\begin{figure}[tb!]
\centering
\includegraphics[width=0.8\linewidth]{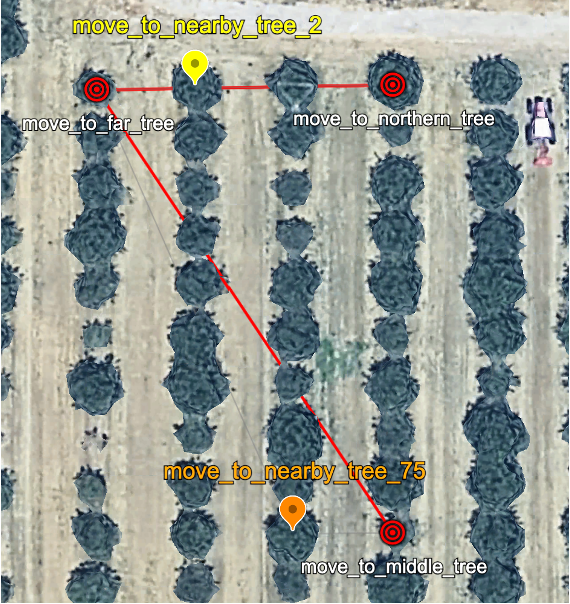}
\caption{Conceptualization of the mission query ``\textit{Find a tree somewhere in the middle of the orchard and measure CO$_2$. If low reading, go to and take a picture of a nearby tree. From there, measure another tree as far away as possible for CO$_2$ and repeat the same process. Once done, take a temperature reading at any one of the northern most trees.}" The yellow and orange trees in the figure represent conditional tasks based on the outcome of the CO$_2$ reading at adjacent trees in red. The red trees are locations that must be visited to satisfy the mission.}
\label{fig:rel_abs}
\vspace{-3mm}
\end{figure}

As compared to the results from \cite{CarpinLLMSubmitted}, the additional step of FV in this paper is much more difficult, as the previous implementation only defined an XSD framework from which the LLM generates the L1 XML mission.
While this provides a syntactic and feasible mission guarantee, this standardization does not provide a guarantee of the semantics represented by the original mission.
We begin with smaller missions from Table \ref{table:results}, missions with fewer tasks and similar to that of Figure \ref{fig:buchi}, and see that FV is successful.
We fundamentally show the ability to decouple mission generation and FV, as in Figure \ref{fig:rel_abs}.

In the case of some of the complex formulas, one example of which is shown in Figures \ref{fig:complex_aut}, we visualize just how complex queries can be.
The figure is without names for states, for legibility, but illustrates
the variety of paths that can emerge during its execution.
In these complex queries, the system demonstrates robustness by being able to overcome conflicts between agents. 
One such case is syntax.  
In the experimentation, many of the retries can be attributed to mismatching parentheses in LTL generation.
This became evident with more complex queries where the generated LTLs would be heavily interconnected.
We note the reduction in these errors with simpler queries.
However, the error feedback loop proves to be enough for the LLM to manage and the generation agents corrected every instance of incorrect syntax within the retry limit.
Another case, previously highlighted in \cite{CarpinLLMSubmitted}, was the case of message ambiguity.
We note, paralleling real-life scenarios, that a sentence can have multiple meanings, and these are the instances where FV is essential to ensure alignment between user intent and automatic mission synthesis.
As an example, one query states, ``\textit{... take 2 CO$_2$ readings of two trees next to each other}." 
This could read as taking a reading between two trees twice, taking two readings of each individual tree, or taking a single reading of each of the trees.
In this example, Spot generated runs of the LTL mission, shared them with the arbiter, and was denied verification due to this  misinterpretation. 
However, showcasing fault tolerance, the LTL generation agent reconsidered its mission interpretation based on this rejection and self-corrected.

\subsection{User Study} \label{sec:user_study}
While the first two sections of Table \ref{table:results} focus on lab-curated missions, we extended the study by reaching out to farm workers and farm 
managers to ask how they might use our system. 
We view these experiments as a stress test on the system to prove that it is capable of generalizing to problems outside the orchard environment. 
The farmers we asked all grow specialty Asian vegetables such as long beans, bok choy, okra, and bittermelon, to name a few. 
All of these crops are row crops with different watering and maintenance schedules. 
Of the roughly 50 example missions we collected from two different farms, we selected those that matched the capabilities of our L1 and L2 system.
Notably, none of the farmers leveraged the conditional logic support provided by the mission planner, but instead used the parameterization capability of the L1 schema for atomic task decomposition.

Since the experiments in \cite{CarpinLLMSubmitted} were configured for an orchard environment, we simply added new robot actions to the existing schema to support the same mobile robot platform.
This increased the available action count from five to nine.
These include relative movement and orientation for planning paths and operating an implement for seeding plants, among others.
Additionally, we added a couple of pieces of context about the environment and robot hardware. 
We used the same farm polygon framework, but mapped it using the boundaries of the collaborating farmer’s field rather than our own test orchard.
Additionally, we included a table mapping the inertial measurement unit's (IMU) raw yaw values into directions.
This gives the planner the ability to orient the robot appropriately for a row crop environment.
Finally, we added a three-line user guide on how the seeder attachment should be operated. 
As an example, the second and third missions in Table \ref{table:results} make spatial references that must be decomposed according to a properly configured L1 schema. 
The full queries read, ``\textit{Plant 40 plants in one row, one yard apart}" and ``\textit{Plant 300 feet long bean, skip one row, plant Chinese bittermelon}" in reference to their two-acre plot. 
Though conditional complexities were not used, we noticed farmers tended to request tasks much more implicitly, making the task equally as difficult for the planner to verify.
As shown by the results from Table \ref{table:results} under the \textit{Farmer} queries, our system was able to take on a completely different environment, from orchard to row crops, with minimal reconfiguration and achieve user-expected results with semantic verification.

\subsection{Limitations} \label{sec:limitations}
Similar to the previous findings in \cite{CarpinLLMSubmitted}, we found that spatial understanding is still an issue for LLMs. 
During our user study, we measured the dimensions of each farmer's field.
When comparing farmer missions with each respective field, we noted that some commands asked the robot to go outside their field to carry out a mission.
As an example, one farmer asked the robot to plant 300 feet of seeds in a row, roughly estimating the length of his field.
However, the true dimension of this field, explicitly supplied as context to the LLM, is $\sim$285 $\times$ $\sim$275 feet.
Meaning, in either direction, the robot could not possibly perform this task without ending up in a potentially dangerous position.
Limitations like these speak to a greater need to not just formally verify action sequences, but also verify meta-constraints that can potentially apply to every mission.
While in our limited user study we note that the participants very infrequently asked for logically complex missions, some of the missions decomposed into longer linear plans of over 20 tasks. 
If broken into several missions, the architecture handles these simply as shown in Table \ref{table:results}. 
However, the architecture is now limited by the number of tokens that the LLM can handle in a single instance. 
We found a correlation between the number of corrected syntax errors and the total size of the decomposed tasks in the mission.
Not done in this paper, one solution is to break up independent sub-missions within a mission query and run them individually so the LLM can evaluate each sub-mission within their token limit.
The most philosophical limitation of the architecture is in choosing which agent to assign blame when no agreement is reached.
As implemented, the architecture operates in series to first approve the XML mission then approve the LTL mission.
However, queries that did not perform correctly every time -- the more complex queries -- often struggled because of a lack of culpability in the initial generation stage.
For example, the architecture starts by generating XML and LTL missions before any verification occurs.
Fundamentally, there must be some agreement on mission structure before it is worthwhile to proceed into arbitration and FV.
As a simple check, so as not to introduce bias into the LLMs, we check that each mission shares the same number of decomposed tasks before moving onto arbitration and SPIN FV.
Because of the well-defined framework provided in XSD, we  assume 
that if there are any errors early in mission decomposition, they come from the LTL generation.
Generally, this assumption stood through all experiments.
Moreover, if we did not pick a single agent to blame, the design faces the problem of flip-flopping LLM answers since they are not directly allowed to communicate to avoid bias.
In more trivial cases where the agents do not immediately match, they can settle on one mission after so many tries because there are only so many plausible interpretations.
The more complex the mission, the more interpretations are allowed.
We experimented with this early on before deciding to isolate one agent at a time. 
First we assume LTL agent fault if there are any discrepancies, then XML agent if FV fails.
We experimented with forms of communication between the agents, but each of them  introduced too much bias.
Ultimately, we opted only to share how many decomposed tasks each agent came up with, which decreases performance but enhances confidence in the FV results. 
\section{Conclusions and Future Work} \label{sec:conclusions}

In this paper, we present an automated pipeline that employs LLMs to decompose a high-level objective specified in natural language to produce mission plans and LTLs for FV that can be used in precision agriculture applications. 
To the best of our knowledge, this is the first implementation of an automated, formally verified, mission planning pipeline.
The solution we proposed enables non-specialists to not only use field robots but to be sure their requests are being satisfied appropriately.
To accomplish this, we introduced a parallel LLM agent with the express task of generating logic for FV, bolstering our claims of a modular system design.
Future research will investigate how this application can be used in distributed robotic systems, applied in potentially infinite or repeating mission requests, and run on edge compute.
Additionally, we wish to continue with our user study to validate the assumptions made during development.  

\bibliographystyle{plain}
\bibliography{report.bib}

\end{document}